\pdfoutput=1

\documentclass[11pt]{article}

\usepackage[]{acl}

\usepackage{times}
\usepackage{latexsym}

\usepackage[T1]{fontenc}

\usepackage[utf8]{inputenc}

\usepackage{microtype}

\usepackage{caption}
\usepackage{graphicx}
\usepackage{subcaption}

\usepackage{multirow}

\title{Event Causality Identification with Causal News Corpus\\ - Shared Task 3, CASE 2022}

\author{Fiona Anting Tan \\
  Institute of Data \\
  Science, National \\
  University of Singapore, \\
  Singapore \\
  \texttt{tan.f@u.nus.edu} \\\And
  Hansi Hettiarachchi \\
  School of Computing and Digital \\
  Technology, Birmingham City \\
  University, United Kingdom \\ \texttt{hansi.hettiarachchi}\\
  \texttt{@mail.bcu.ac.uk} \\\And
  Ali Hürriyetoğlu \\
  KNAW Humanities \\
  Cluster DHLab, \\
  The Netherlands \\ 
  \texttt{ali.hurriyetoglu}\\
  \texttt{@dh.huc.knaw.nl}\\\AND
  Tommaso Caselli \\
  CLCG, \\
  University of Groningen, \\
  The Netherlands \\
  \texttt{t.caselli@rug.nl} \\\And
  Onur Uca \\
  Department of Sociology, \\
  Mersin University, \\
  Turkey \\ 
  \texttt{onuruca@mersin.edu.tr} \\\And
  Farhana Ferdousi Liza \\
  School of Computing Sciences, \\
  University of East Anglia, \\
  United Kingdom \\
  \texttt{F.Liza@uea.ac.uk} \\\AND
  Nelleke Oostdijk \\
  Centre for Language Studies, \\
  Radboud University, The Netherlands \\
  \texttt{nelleke.oostdijk@ru.nl} \\
  }

\begin{document}
\maketitle
\begin{abstract}
The Event Causality Identification Shared Task of CASE 2022 involved two subtasks working on the Causal News Corpus. Subtask 1 required participants to predict if a sentence contains a causal relation or not. This is a supervised binary classification task. Subtask 2 required participants to identify the Cause, Effect and Signal spans per causal sentence. This could be seen as a supervised sequence labeling task. For both subtasks, participants uploaded their predictions for a held-out test set, and ranking was done based on binary F1 and macro F1 scores for Subtask 1 and 2, respectively. This paper summarizes the work of the 17 teams that submitted their results to our competition and 12 system description papers that were received. The best F1 scores achieved for Subtask 1 and 2 were 86.19\% and 54.15\%, respectively. All the top-performing approaches involved pre-trained language models fine-tuned to the targeted task. We further discuss these approaches and analyze errors across participants' systems in this paper.
\end{abstract}

\section{Introduction}
A causal relation represents a semantic relationship between a Cause argument and an Effect argument, in which the occurrence of the Cause leads to the occurrence of the Effect \citep{DBLP:journals/rcs/BarikMO16}. Extracting causal information from text has many downstream natural language processing (NLP) applications, for summarization and prediction \cite{DBLP:conf/www/RadinskyDM12, DBLP:conf/wsdm/RadinskyH13, izumi-etal-2021-economic, hashimoto-etal-2014-toward}, question answering \cite{dalal-etal-2021-enhancing, ijcai2019-0695, stasaski-etal-2021-automatically}, inference and understanding \cite{DBLP:journals/tacl/JoBRH21, dunietz-etal-2020-test}. 


However, data for causal text mining is limited \citep{asghar2016automatic, xu-etal-2020-review, DBLP:journals/kais/YangHP22, tan-etal-2021-causal, tan-EtAl:2022:LREC}. There are also not many benchmarks to allow for fair model comparisons \cite{asghar2016automatic}. Therefore, in this paper, we continue our efforts with the creation of the Causal News Corpus (CNC). CNC is a corpus of news articles annotated with causal information suitable for causal text mining. Additionally, we introduce a shared task to promote modelling for two causal text mining tasks: (1) Causal Event Classification and (2) Cause-Effect-Signal Span Detection. Figure \ref{fig:examples} provides examples from the CNC in this shared task. To our knowledge, we are the first dedicated causal text mining dataset and benchmark to include signal span detection as an objective.

\begin{figure}[h]
  \centering
  \includegraphics[scale=0.13]{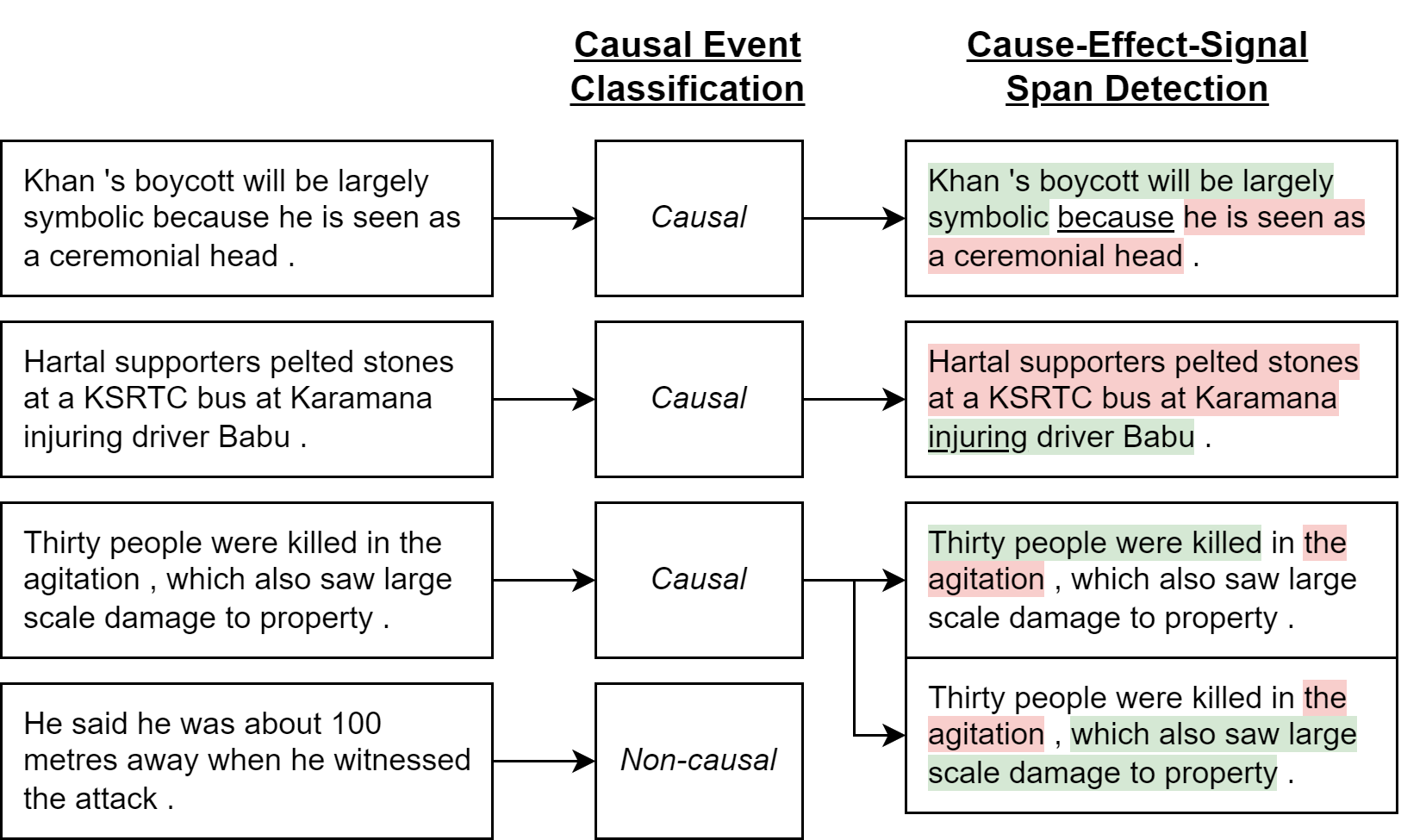}
  \caption{Examples from the CNC for the two subtasks. Cause spans are indicated by Pink, while Effect spans are indicated by Green. Signals, if present, are underlined.}
\end{figure}\label{fig:examples}

The rest of the paper is organized as follows: Section \ref{sec:related} presents literature on event causality datasets. Section \ref{sec:dataset} describes the dataset and annotation of the corpus. Section \ref{sec:task} formally introduces the two subtasks for the shared task. Section \ref{sec:eval_competition} describes the evaluation metrics and competition set-up. Subsequently, Section \ref{sec:participant} summarizes the methods used by participants during the competition, while Section \ref{sec:analysis} analyzes the participants' submissions. Finally, Section \ref{sec:conclusion} concludes this paper.

\section{Related Work}
\label{sec:related}

In many papers about Event Causality Identification (ECI) \citep{gao-etal-2019-modeling, zuo-etal-2021-learnda, cao-etal-2021-knowledge, zuo-etal-2021-improving, zuo-etal-2020-knowdis}, the two datasets used for benchmarking are CausalTimeBank \cite{mirza-etal-2014-annotating,mirza-tonelli-2014-analysis} and EventStoryLine \cite{caselli-vossen-2017-event}. These datasets are unsuitable for span detection since their arguments are event headwords only.

There are two other efforts that intentionally introduce datasets for benchmarking causal text mining systems. FinCausal \citep{mariko-etal-2021-financial, mariko-etal-2020-financial} is a recurring shared task held within the FinNLP workshop focusing on financial news. In the first subtask, participants also aim to identify if sentences contain causal relations. In the second subtask, participants to identify the Cause and Effect spans in the causal sentences. UniCausal \cite{unicausal}\footnote{\url{https://github.com/tanfiona/UniCausal}} is an open-source repository for causal text mining that has consolidated six corpora for three causal text mining tasks. The six corpora included in UniCausal are: AltLex \cite{hidey-mckeown-2016-identifying}, BECAUSE 2.0 \cite{dunietz-etal-2017-corpus}, CausalTimeBank \cite{mirza-etal-2014-annotating,mirza-tonelli-2014-analysis}, EventStoryLine V1.0 \cite{caselli-vossen-2017-event}, Penn Discourse Treebank V3.0 \cite{webber2019penn}, and SemEval 2010 Task 8 \cite{hendrickx-etal-2010-semeval}. The three tasks are: Causal Sentence Classification, Causal Pair Classification and Cause-Effect Span Detection.

Similar to FinCausal and UniCausal, we included a signal span detection objective. Our annotation guidelines differ slightly, in that our arguments must contain events, and the spans are annotated in a manner that is minimally sufficient. In general, we notice that spans from FinCausal are much longer. Spans from UniCausal depend on the original data source. 

Additionally, for Cause-Effect Span Detection in FinCausal, their approach to handle multiple causal relations per unique sentence was to include index numbers at the start of each sentence to differentiate the Cause-Effect predictions. This approach is problematic because (1) it leaks information that the sentence contains multiple causal relations to the model, and (2) predictions that are submitted in a different order from the true labels are unnecessarily penalised. Therefore, we differ from FinCausal when evaluating multiple causal relations in span detection since we group relations by its sentence index. This is described further in Section \ref{ssec:eval}.

\section{Dataset}
\label{sec:dataset}

\subsection{Data Collection}
\label{ssec:data_collection}

Our shared task worked with the Causal News Corpus (CNC) \citep{tan-EtAl:2022:LREC}\footnote{\url{https://github.com/tanfiona/CausalNewsCorpus}}, which consists of 869 news documents and 3,559 English sentences, annotated with causal information. CNC builds on the randomly sampled articles~\cite{doi:10.1177/00027642211021630} from multiple sources and periods featured~\cite{10.1162/dint_a_00092} in a series of workshops directed at mining socio-political events from news articles \cite{hurriyetoglu-etal-2020-automated,aespen-2020-automated, hurriyetoglu-etal-2021-multilingual,hurriyetoglu-etal-2021-challenges,case-2021-challenges}. CNC follows the train-test split of the original data source, with 3,248 training and 311 test examples. Later, we further split and randomly sampled 10\% of the original training set to obtain the development set. Later, Table \ref{tab:subtask1_data} presents the sentence counts per data split.

\subsection{Annotation}
\label{ssec:annotation}

\subsubsection{Guidelines}
For more information on our annotation guidelines, please refer to our annotation manual\footnote{Available under the "documentation" folder of CNC's Github repository.}.

\paragraph{Subtask 1}
In CNC, sentences were labeled as \emph{Causal} or \emph{Non-causal}, where the presence of causality indicates that “one argument provides the reason, explanation or justification for the situation described by the other” \citep{webber2019penn}. Our sentences had to contain at least a pair of events, defined as ``things that happen or occur, or states that are valid" \citep{sauri2006timeml}. These annotations correspond to the target labels for Subtask 1, Causal Event Classification.

\paragraph{Subtask 2}
For \emph{Causal} sentences, the words corresponding to the Cause-Effect-Signal spans of a causal relation were also marked. These annotations correspond to the target labels for Subtask 2, Cause-Effect-Signal Span Detection. However, at the current stage of writing, only a small subset of our data contains annotated spans. Span annotations are an on-going effort.

A Cause is a reason, explanation or justification that led to an Effect. We defined a Cause or Effect span as a continuous set of words sufficient for the interpretation of the causal relation meaning. This means that any context modifying or describing the argument relevant to the causal relation was included. Each Cause or Effect span must contain an event, where an event is defined as a situation that `happen or occur', or predicates that `describe states or circumstances in which something obtains or holds true' \citep{DBLP:conf/ndqa/PustejovskyCISGSKR03}. 

Signals are words that help to identify the structure of the discourse. In our case, signals highlight the relationship between the Cause and Effect.

\subsubsection{Annotation Tool}
We used the WebAnno tool \citep{eckart-de-castilho-etal-2016-web} to conduct our annotation process.

\paragraph{Subtask 1}
Annotation at the sequence level was relatively straightforward, where annotators selected ``Yes" or ``No" labels for each sentence.

\paragraph{Subtask 2}

\begin{figure*}[h]
  \centering
  \includegraphics[scale=0.4]{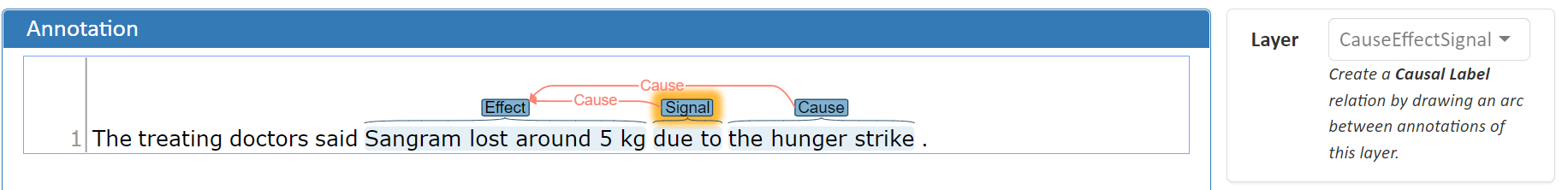}
  \caption{Screenshot of the annotation tool used to mark Cause-Effect-Signal spans.}
  \label{fig:anno_tool}
\end{figure*}

Annotators first marked the Cause span, Effect span, and Signal span. Subsequently, they linked the spans together by pointing Cause to Effect and Signal to Effect. An illustration is provided in Figure \ref{fig:anno_tool}. Annotations were then downloaded and sent through checking scripts on Python to identify if there were any avoidable human errors. For example, if missing links (E.g. An Effect has no Cause) or invalid links (E.g. An Effect points to Effect) were present, and an error report was then sent to annotators for them to consider correcting their annotations.


\subsubsection{Annotation Process \& Curation}
Five annotators were involved and independently annotated for both subtasks across the span of a few months. For each round of annotations, annotators were presented with a subset of the dataset. After each round, the curator consolidated the final annotations as follows:

\paragraph{Subtask 1}
The majority voted label was retained as the final label. Every example in the final corpus was annotated by at least two annotators. The curator has the final vote if there are ties, or if only one annotation is present. Further details are available in the CNC paper \citep{tan-EtAl:2022:LREC}.

\paragraph{Subtask 2} 
There was no straightforward way to take a majority label for span annotations. Therefore, our approach was that the curator took into account the spans highlighted by the annotators and decided on the final selection. 

After each annotation round, the final span annotations were made available for annotators to review and discuss.

\subsubsection{Summary Statistics}

\paragraph{Inter-annotator Agreement}

\begin{table}[]
\centering
 \resizebox{0.9\columnwidth}{!}{
\begin{tabular}{llcccc}\hline
 & Train & Dev & Test & Total \\\hline
K-Alpha & 34.42 & 29.77 & 48.55 & 34.99\\\hline
\end{tabular}
}
\caption{Subtask 1 Inter-annotator Agreement Scores. Reported in percentages.}\label{tab:subtask1_agreement}
\end{table}
\begin{table}[]
\centering
 \resizebox{0.95\columnwidth}{!}{
\begin{tabular}{p{10mm}lccc}\hline
Metric & Span & Train+Dev & Test & Total \\\hline
\multirow{4}{10mm}{Exact Match} & Cause & 30.57 & 15.11 & 23.88 \\
 & Effect & 36.30 & 19.86 & 29.19 \\
 & Signal & 27.92 & 29.21 & 28.48 \\\cline{2-5}
 & Total & 7.84 & 5.81 & 6.96 \\\hline
\multirow{4}{10mm}{One-Side Bound} & Cause & 57.55 & 39.86 & 49.90 \\
 & Effect & 60.90 & 45.42 & 54.21 \\
 & Signal & 31.93 & 32.96 & 32.37 \\\cline{2-5}
 & Total & 24.05 & 22.25 & 23.27 \\\hline
\multirow{4}{10mm}{Token Overlap} & Cause & 63.65 & 49.18 & 57.39 \\
 & Effect & 64.66 & 49.88 & 58.27 \\
 & Signal & 32.09 & 33.15 & 32.55 \\\cline{2-5}
 & Total & 26.94 & 27.78 & 27.31 \\\hline
\multirow{4}{10mm}{K-Alpha} & Cause & 46.36 & 42.51 & 44.32 \\
 & Effect & 57.18 & 41.89 & 49.89 \\
 & Signal & 29.30 & 23.42 & 27.08 \\\cline{2-5}
 & Total & 50.90 & 41.54 & 46.27\\\hline
\end{tabular}
}
\caption{Subtask 2 Inter-annotator Agreement Scores. Reported in percentages (\%).}\label{tab:subtask2_agreement}
\end{table}

For Subtask 1, scores are reflected in Table \ref{tab:subtask1_agreement}. Also reported in \citet{tan-EtAl:2022:LREC}, overall, the dataset has a Krippendorff's Alpha (K-Alpha) agreement score of 34.99\%.

For Subtask 2, the agreement metrics used were Exact Match (EM), Token Overlap (TO), One-Side Bound (OSB), and K-Alpha. Scores are presented in Table \ref{tab:subtask2_agreement}. Overall, the dataset had agreement scores of 6.96\% EM, 23.27\% OSB, 27.31\% TO, and 46.27\% K-Alpha. Since OSB and TO are relaxed span evaluation metrics \citep{DBLP:conf/sigir/LeeS19}, they are naturally much higher than EM, which is a strict metric. How the metrics were calculated is described in the Appendix Section \ref{ssec:agreement}.

\paragraph{Shared Task Data}
The summary statistics for Subtask 1 and 2 are available in Tables \ref{tab:subtask1_data} and \ref{tab:subtask2_data} respectively. 

It is worth noting that for Subtask 2, the test set contained sentences that were much longer than those in the training sets. This is because we were annotating the shorter sentences first based on annotators' feedback that working with shorter sentences at the beginning helps them to familiarise themselves with the annotation rules. Since there were more sentences in the training set, the training set naturally also had more short sentences for us to annotate first. Once we are done with span annotations, the average number of words for Subtask 2 should tally with the causal sentences of Subtask 1, shown earlier in Table \ref{tab:subtask1_data}.

\begin{table}[]
\centering
 \resizebox{1\columnwidth}{!}{
\begin{tabular}{p{7mm}p{18mm}cccc}\hline
Stat. & Label & Train & Dev & Test & Total \\\hline
\multirow{3}{7mm}{\# Sent-ences} & \emph{Causal} & 1603 & 178 & 176 & 1957 \\
 & \emph{Non-causal} & 1322 & 145 & 135 & 1602 \\\cline{2-6}
 & Total & 2925 & 323 & 311 & 3559 \\\hline
\multirow{3}{7mm}{Avg. \# words} & \emph{Causal} & 35.48 & 36.86 & 41.27 & 36.13 \\
 & \emph{Non-causal} & 27.34 & 27.35 & 30.25 & 27.59 \\\cline{2-6}
 & Total & 31.80 & 32.59 & 36.49 & 32.28\\\hline
\end{tabular}
}
\caption{Subtask 1 Data Summary Statistics.}\label{tab:subtask1_data}
\end{table}
\begin{table}[]
\centering
 \resizebox{1\columnwidth}{!}{
\begin{tabular}{lcccc}\hline
Stat. & Train & Dev & Test & Total \\\hline
\# Sentences & 160 & 15 & 89 & 264 \\
\# Relations & 183 & 18 & 119 & 320 \\\cline{2-5}
Avg. rels/sent & 1.14 & 1.20 & 1.34 & 1.21 \\\hline
Avg. \# words & 17.21 & 16.13 & 28.45 & 20.94 \\
\hspace{3mm} Cause & 6.52 & 7.28 & 12.76 & 8.89 \\
\hspace{3mm} Effect & 7.80 & 6.44 & 10.20 & 8.62 \\
\hspace{3mm} Signal & 1.55 & 1.60 & 1.36 & 1.47 \\\hline
Avg \# signals/rel & 0.67 & 0.56 & 0.82 & 0.72 \\
Prop. of rels w/ signals & 0.64 & 0.56 & 0.76 & 0.68\\\hline
\end{tabular}
}
\caption{Subtask 2 Data Summary Statistics.}\label{tab:subtask2_data}
\end{table}

\section{Task Description}
\label{sec:task}
The shared task is comprised of two subtasks related to Event Causality Identification. The objective of each task is described in detail as follows:

\subsection{Subtask 1: Causal Event Classification}
The objective of this task is to classify whether an event sentence contains any cause-effect meaning. Systems had to predict \emph{Causal} or \emph{Non-causal} labels per test sentence. An event sentence was defined to be \emph{Causal} if it contains at least one causal relation.

\subsection{Subtask 2: Cause-Effect-Signal Span Detection}
The objective of this task is the detection of the consecutive spans relevant to a \emph{Causal} relation. There are three types of spans involved in a \emph{Causal} relation: The \emph{Cause} span refers to words that describe the event that triggers another \emph{Effect} event. The \emph{Effect} span refers to words that describe the resulting event arising from a \emph{Cause} event. \emph{Signals} are optionally present, and are words that explicitly indicate a \emph{Causal} relation is present. In our dataset, multiple \emph{Causal} relations can exist in a sentence, and participants have to identify all of them.

\section{Evaluation \& Competition}
\label{sec:eval_competition}

\subsection{Evaluation Metrics}
\label{ssec:eval}

\subsubsection{Subtask 1}
We evaluated participants' predictions using Accuracy (Acc), Precision (P), Recall (R), F1, and Matthews Correlation Coefficient (MCC) scores. 

\subsubsection{Subtask 2}
Following previous evaluation metrics for Cause-Effect Span Detection \citep{mariko-etal-2020-financial, mariko-etal-2021-financial} and text chunking \citep{tjong-kim-sang-buchholz-2000-introduction}, we assessed predictions using Macro P, R and F1 metrics.

Participants uploaded sentences with Cause-Effect-Signal spans marked directly in the text using \texttt{ARG0}, \texttt{ARG1} and \texttt{SIG} start and end boundary markers. We converted these marked sentences into two white-space tokenized sequences, one corresponding to the token labels for Cause and Effect, and another corresponding to the token labels for Signals. We used the token classification evaluation scheme from \texttt{seqeval} \citep{seqeval, ramshaw-marcus-1995-text}\footnote{\url{https://github.com/chakki-works/seqeval}} provided through Huggingface \citep{wolf-etal-2020-transformers}\footnote{\url{https://huggingface.co/spaces/evaluate-metric/seqeval}}.

Evaluation was conducted at the relation level. In other words, examples with multiple causal relations were unpacked and each relation contributed equally to the final score.

\paragraph{Handling multiple relations} 

Since one input sequence can return multiple causal relations, we adjusted the evaluation code to automatically extract the combination that results in the best F1 score. As such, participants could submit multiple Cause-Effect-Signal span predictions per input sequence in any order. An illustration is provided in Figure \ref{fig:st2_evaluation}. 

In evaluation, we only compare with the number of causal relations that the true label has. Let the number of predicted relations be $n_p$, and the number of actual relations be $n_a$. Our evaluation script does the following:

\begin{itemize}
    \item If the number of predicted relations exceeds the number of actual relations ($n_p>n_a$), we kept only the first $n_a$ predictions.
    \item If the number of predicted relations is less than the number of actual relations ($n_p<n_a$), the missing $n_a-n_p, n_a-n_p+1, ... , n_a$ relation predictions were represented by tokens that all correspond to the Other (\texttt{O}) label.
\end{itemize}

\begin{figure*}[h]
  \centering
  \includegraphics[scale=0.17]{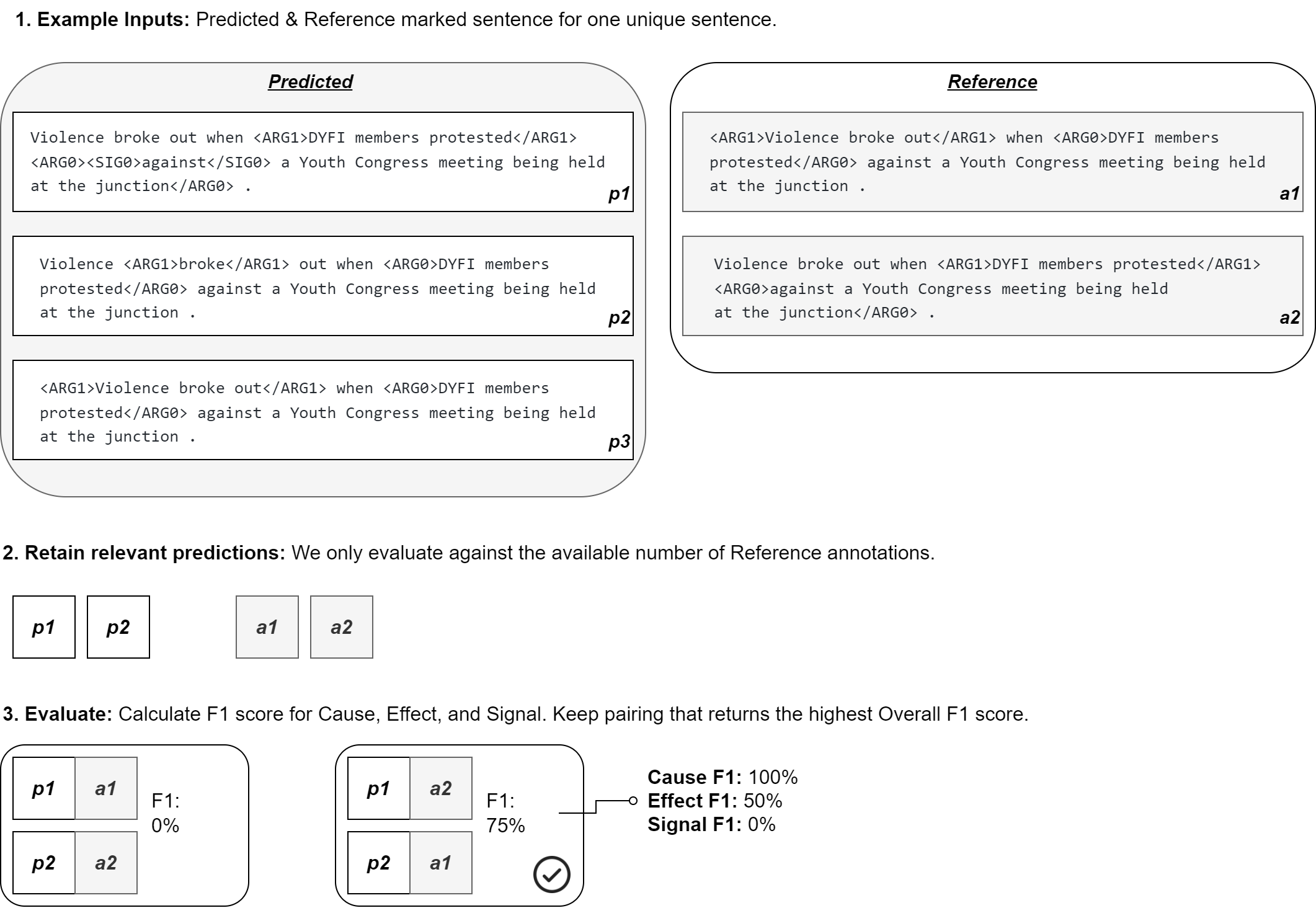}
  \caption{Illustration of how we process multi-relation examples for sequence evaluation.}
\end{figure*}\label{fig:st2_evaluation}

\subsection{Baseline}
For Subtask 1, we duplicated the BERT \cite{devlin-etal-2019-bert} and LSTM \cite{hochreiter1997long} baselines from our previous work \citep{tan-EtAl:2022:LREC} that achieved F1 scores of 81.20\% and 78.22\% respectively.

For Subtask 2, a random baseline\footnote{\url{https://github.com/tanfiona/CausalNewsCorpus/blob/master/random_st2.py}} was created for reference. This baseline first randomly identifies start positions for Cause and Effect spans, and then identifies end positions for these spans with a linearly increasing probability as we move away from the start location in order to reflect our preference for longer spans. We also randomly predicted words to be signals with a 10\% chance. The baseline F1 score was 0.45\%.

\subsection{Competition Set-up}
We used the Codalab website to host our competition.\footnote{The competition page is at \url{https://codalab.lisn.upsaclay.fr/competitions/2299}. The additional scoring page is at \url{https://codalab.lisn.upsaclay.fr/competitions/7046}.}

\paragraph{Registration} 37 participants requested to participate on the Codalab page. However, we required participants to email us some personal details (Name, Institution and Email) to avoid teams from creating multiple accounts to cheat. Subsequently, 29 participants were successfully registered, but only 17 accounts participated by uploading predictions.

\paragraph{Trial and Test Periods} The trial period started on April 15, 2022 and the validation labels were released on August 01, 2022. Participants could upload any number of submissions against the validation set, and they could also submit predictions for the validation set at any point in time. The main purpose of this setting is for participants to familiarise themselves with the Codalab platform. 

The test period started on August 01, 2022 and ended on August 31, 2022. Each participant was allowed only 5 submissions to prevent participants from over-fitting to the test set. After the competition ended, an additional scoring page was created,\footnote{The additional scoring page is at \url{https://codalab.lisn.upsaclay.fr/competitions/7046}.} where participants could upload one prediction a day to generate more scores for their description papers. Any scores from this additional scoring page is not included into the final leaderboard.

For both subtasks, models were ranked based on F1 performance on the competition test set.

\section{Participant Systems}
\label{sec:participant}

\subsection{Overview}
13 participants successfully submitted scores to Subtask 1 while only 4 successfully submitted scores to Subtask 2 during test period. Table \ref{tab:subtask1_ldbrd} and \ref{tab:subtask2_ldbrd} reflects the leaderboard for Subtask 1 and 2 respectively for evaluation metrics described earlier in Section \ref{ssec:eval}. For Subtask 2, we further provided the performance for each span type (i.e., Cause, Effect and Signal). 

\begin{table*}[]
\centering
 \resizebox{0.9\textwidth}{!}{
\begin{tabular}{cll|ccccc}\hline
Rank & Team Name & Codalab Username & R & P & F1 & Acc & MCC\\\hline
1 & CSECU-DSG \citep{23} & csecudsg & 88.64 & \textbf{83.87} & \textbf{86.19} & \textbf{83.92} & \textbf{67.14} \\
2 & ARGUABLY \citep{24} & guneetsk99 & \textbf{91.48} & 81.31 & 86.10 & 83.28 & 66.02 \\
3 & LTRC \citep{10} & hiranmai & 88.64 & 82.11 & 85.25 & 82.64 & 64.51 \\
4 & NLP4ITF \citep{6} & pogs2022 & 88.07 & 82.45 & 85.16 & 82.64 & 64.49 \\
5 & IDIAPers \citep{13} & msingh & 87.50 & 82.80 & 85.08 & 82.64 & 64.49 \\
6 & NoisyAnnot \citep{15} & thearkamitra & 88.07 & 82.01 & 84.93 & 82.32 & 63.83 \\
7 & SNU-Causality Lab \citep{9} & JuHyeon\_Kim & 90.34 & 79.50 & 84.57 & 81.35 & 62.04 \\
8 & LXPER AI Research & brucewlee & 86.36 & 82.61 & 84.44 & 81.99 & 63.18 \\
9 & 1Cademy \citep{17} & nika & 86.36 & 81.72 & 83.98 & 81.35 & 61.85 \\
10 & - & quynhanh & 85.80 & 79.06 & 82.29 & 79.10 & 57.19 \\
11 & BERT Baseline \citep{tan-EtAl:2022:LREC} & tanfiona & 84.66 & 78.01 & 81.20 & 77.81 & 54.52 \\
12 & GGNN \citep{16} & PaulTrust & 88.07 & 74.88 & 80.94 & 76.53 & 52.05 \\
13 & LSTM Basline \citep{tan-EtAl:2022:LREC} & hansih & 84.66 & 72.68 & 78.22 & 73.31 & 45.15 \\
14 & Innovators & lapardnemihk9989 & 78.98 & 72.02 & 75.34 & 70.74 & 39.81 \\
15 & - & necva & 81.25 & 59.09 & 68.42 & 57.56 & 9.44\\\hline
\end{tabular}
 }
\caption{Subtask 1 Leaderboard. Ranked by Binary F1. All scores are reported in percentages (\%). Highest score per column is in bold.}\label{tab:subtask1_ldbrd}
\end{table*}
\begin{table*}[]
\centering
 \resizebox{1\textwidth}{!}{
\begin{tabular}{p{3mm}lp{14mm}|cccc|ccc|ccc|ccc}\hline
\multirow{2}{3mm}{Ra-nk} & \multicolumn{1}{c}{\multirow{2}{*}{Team Name}} & \multicolumn{1}{c|}{\multirow{2}{14mm}{Codalab Username}} & \multicolumn{4}{c}{Overall} & \multicolumn{3}{|c|}{Cause (n=119)} & \multicolumn{3}{|c|}{Effect (n=119)} & \multicolumn{3}{|c}{Signal (n=98)} \\
 & \multicolumn{1}{c}{} & \multicolumn{1}{c|}{} & R & P & F1 & Acc & R & P & F1 & R & P & F1 & R & P & F1 \\\hline
1 & 1Cademy \citep{18} & gezhang & \textbf{53.87} & 55.09 & \textbf{54.15} & \textbf{43.15} & \textbf{55.46} & \textbf{57.98} & \textbf{56.47} & \textbf{55.46} & 57.14 & \textbf{56.13} & 50.00 & 49.09 & 48.92 \\
2 & IDIAPers \citep{14} & msingh & 47.62 & 51.21 & 48.75 & 40.83 & 45.38 & 45.38 & 45.38 & 42.86 & 42.86 & 42.86 & \textbf{56.12} & \textbf{68.44} & \textbf{60.01} \\
3 & SPOCK \citep{22} & spock & 43.75 & \textbf{57.62} & 47.48 & 36.87 & 37.82 & 49.19 & 41.40 & 39.50 & \textbf{59.66} & 46.29 & 56.12 & 65.39 & 56.32 \\
4 & LTRC \citep{10} & hiranmai & 5.65 & 2.34 & 3.23 & 33.03 & 2.52 & 1.10 & 1.53 & 13.45 & 5.51 & 7.60 & 0.00 & 0.00 & 0.00 \\
5 & Random Baseline & tanfiona & 0.30 & 0.89 & 0.45 & 21.94 & 0.84 & 2.52 & 1.26 & 0.00 & 0.00 & 0.00 & 0.00 & 0.00 & 0.00\\\hline
\end{tabular}
 }
\caption{Subtask 2 Leaderboard. Ranked by Overall Macro F1. All scores are reported in percentages (\%). Highest score per column is in bold.}\label{tab:subtask2_ldbrd}
\end{table*}

For Subtask 1, the top performing team was CSECU-DSG \citep{23}, scoring 86.19\% F1. CSECU-DSG also topped the charts for P, Acc, and MCC scores. Team ARGUABLY \citep{24} followed closely after, with 86.10\% F1 score and a high recall score of 91.48\%. Both methods fine-tuned SOTA pre-trained BERT variants (RoBERTA \citep{DBLP:journals/corr/abs-1907-11692} and DeBERTa \citep{DBLP:conf/iclr/HeLGC21}) to the classification task.

For Subtask 2, the top performing team was 1Cademy \citep{18}, scoring 54.15\% F1. Team IDIAPers \citep{14} and SPOCK \citep{22} followed closely after, with 48.75\% and 47.48\% F1 scores respectively. Each team approached the span detection task in a different way: 1Cademy treated the task as a reading comprehension challenge and predicted start and end boundaries of the spans. IDIAPers treated the task as a decoding challenge, while SPOCK generated and classified candidate spans. All participants used pre-trained models in their frameworks.

\subsection{Methods}
Each teams' systems are summarized below, sorted according to their leaderboard ranking.

\subsubsection{Subtask 1}
\paragraph{CSECU-DSG} \citep{23} proposed a way to unify predictions obtained from two neural network models, by combining the prediction scores generated from each model using a weighted arithmetic mean. The two models used were, Twitter RoBERTa and RoBERTa-base, and each was attached to a linear layer to predict the causal labels. The weights per model were 0.4 and 0.6 respectively, selected through experiments on training data. Their findings on the test set showed that the fused model achieves higher P, R, and F1 score than each model alone, and their approach clinched the top place during the competition.

\paragraph{ARGUABLY} \citep{24} proposed using sentence-level data augmentation to fine-tune language models (LMs). They involved contextualised word embeddings of DistilBERT \citep{DBLP:journals/corr/abs-1910-01108} to construct new data. As for the LMs, DeBERTa and dual cross attention RoBERTa models have been experimented with. According to the results, the DeBERTa model fine-tuned on augmented data outperformed the unaugmented DeBERTa model and RoBERTa models.

\paragraph{LTRC} \citep{10} used various transformers-based language models followed by a classification head. The pre-trained models explored by them were: BART-large \citep{DBLP:conf/acl/LewisLGGMLSZ20}, RoBERTa-base+Linear Layer, RoBERTa-large+Linear Layer, RoBERTa-base+Adapter and RoBERTa-large+Adapter. Their best model slightly beats the baseline scores on the development set.

\paragraph{NLP4ITF} \citep{6} proposed building a RoBERTa model with linguistic features. They mainly involved named entities (NE) and cause-effect-signal (CES) spans from Subtask 2 to incorporate linguistic features with the input text. Based on their findings, the model trained with the PER (person) NE class with CES, achieved the best results, outperforming the RoBERTa baseline (model trained on data with no linguistic features).

\paragraph{IDIAPers} \citep{13} proposed a prompt-based approach for fine-tuning LMs in which the classification task is modeled as a masked language modeling problem (MLM). This approach allows LMs natively pre-trained on MLM problems, like RoBERTa, to directly generate textual responses to domain-specific prompts. This approach allow the model to be trained in a few-shot configuration, keeping most of available data for measuring the generalization power the model. The best-performing model was trained with only 256 instances per class and yet was able to obtain the second-best precision and third-best accuracy.

\paragraph{NoisyAnnot} \citep{15} proposed fine-tuning different LMs with customised cross-entropy loss functions that exploit annotation information such as the number of annotators and their agreement. They used several language models including BERT, RoBERTa and XLNET models and showed that the involvement of annotation information improves the model performance.

\paragraph{SNU-Causality Lab} \citep{9} proposed fine-tuning an ELECTRA model using the CNC dataset and augmented data. They followed two approaches for data augmentation: (1) concatenating SemEval-2010 to CNC and (2) generating new samples using POS tagging. With the POS tagging-based approach they mainly targeted replacing causality irrelevant words with POS tags, to generate more data while preserving the causality relevant information in the original dataset.

\paragraph{1Cademy} \citep{17} experimented with self-training to generate more sequence classification examples from unlabeled Wikipedia sentences. They experimented with three pretrained models (BERT, RoBERTa and ELECTRA), and also experimented with three ratios of positive to negative self-labeled examples (1:3, 1:1, 3:1). Their experiments showed that including self-labeled data during training always returns higher F1 scores. Their best model during test time was the RoBERTa-based model with 1:1 self-training ratio, which surpassed the competition baseline scores.


\paragraph{GGNN} \citep{16} injected word embeddings into a Gated Graph Neural Network (GGNN), which were attached to a RNN decoder to predict the sequence label. Two word embeddings were explored: Word2Vec and BERT. Their BERT+GGNN combination outperforms the BERT baseline provided during the competition for both the development and test sets for P, F1 and Acc.

\subsubsection{Subtask 2}

\paragraph{1Cademy} \citep{18} approached this task in a reading comprehension manner, and created a baseline BERT-based neural network that predicted the start and end positions of each Cause, Effect, and Signal span. They introduced beam-search methods (BSS) as post-processing constraints suited to the task. They also introduced a signal classifier that detects if a Signal exists in the sequence or not via a joint model (JS) or a separate model (ES). Additionally, BART was fine-tuned for paraphrasing to re-write Cause and Effect phrases within each sentence for data augmentation (DA). In the end, their best model is a combination of Baseline+BSS+ES+DA method, where the DA generated 3 new phrases per span. This model achieved F1 score of 54.15\% on the test set, clinching the top place during the competition.

\paragraph{IDIAPers} \citep{14} approached the task in an encoder-decoder framework. They conditioned the T5 language model three times per example to generate up to four causal relations per example. In each round, given the history of a sentence, the model generates Cause, followed by Effect, and then Signal. This model is their vanilla model known as T5-CES. History refers to the input sequence with any annotated spans from the previous round, if applicable. In experiments, they also explored (1) variants involving a version without historical annotations, (2) T5-large pre-trained model, and (3) changing the order of generation to be Effect, Cause then Signal. Their best model on the test set (T5-CES) achieved 48.8 F1 score, coming in second in the competition.

\paragraph{SPOCK} \citep{22} designed two separate frameworks for the span detection task, span-based modelling and token classification. Both approaches far exceed the random baseline provided by the organizers during the competition period. Their span-based modelling approach achieved an F1 score of 47.48\%, ranking third in the competition. This model classifies a list of candidate spans to a Cause, Effect, Signal or None label. The candidate spans are generated by considering all possible spans up to a maximum length. The model receives inputs comprising a CLS token embedding, concatenated with a width embedding, plus the span embedding representation itself. To select the final Cause-Effect-Signal span, spans below a certain threshold are removed, and then the span with the highest probability for that label is retained.

\paragraph{LTRC} \citep{10} approached the task as a token classification task, and designed a BERT-based IOB predicting model alongside some heuristics adjusted for the task. Their approach slightly beats the baseline scores on the development set.

\section{Analysis \& Discussion}
\label{sec:analysis}

\begin{figure*}[ht]
\begin{subfigure}{.5\textwidth}
  \centering
  \includegraphics[width=.7\linewidth]{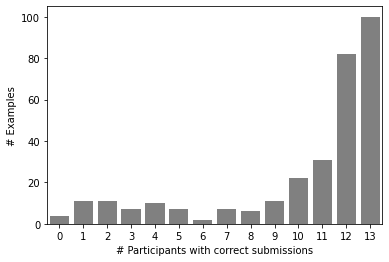}
  \caption{Subtask 1}
\end{subfigure}
\begin{subfigure}{.5\textwidth}
  \centering
  \includegraphics[width=.7\linewidth]{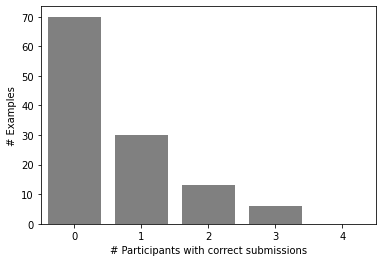}
  \caption{Subtask 2}
\end{subfigure}
\caption{Histogram of exact matches.}
\label{fig:errors}
\end{figure*}

\subsection{Trends}
Consistent with NLP trends, pre-trained language models are popular and employed by all teams and for both subtasks. 

For Subtask 1, teams found novel ways to improve from the BERT and LSTM baseline by combining multiple models, adding linguistic features, incorporating additional neural network layers, and working with augmented data.

For Subtask 2, there is a wide variation in framing the task. Teams approached it as a reading comprehension, encoder-decoder, candidate span classification and token classification task. Additionally, there are two constraints that models had to accommodate: (1) The task involves predicting multiple causal relations per input sentence, and (2) Not all causal relations have a signal span. The top three teams carefully adjust their models to work with the two constraints. For (1), IDIAPers predicted different relations using rounds while incorporating the predicted annotations of the previous round. For (2), 1Cademy included a separate classification task, while IDIAPers and SPOCK permitted "empty" or "None" span predictions. Interestingly, the F1 score for signals is highest for IDIAPers, suggesting merits to predict signal spans in a manner that includes Cause and Effect predictions as inputs. 

\subsection{Participation}




\begin{table}[]
\centering
\begin{tabular}{l|cc|c}\hline
Subtask & Finished & Failed & Total \\\hline
Subtask 1 & 58 & 8 & 66 \\
Subtask 2 & 12 & 24 & 36\\\hline
\end{tabular}
\caption{Number of submissions received for test set.}\label{tab:submission_count}
\end{table}

More submissions were received for Subtask 1 than for Subtask 2, as shown in Table \ref{tab:submission_count}. Unsurprisingly, there is a high proportion of failed submissions in Subtask 2. Since Subtask 2 requires specific formatting of argument markings and compiling of multiple predictions into a list, it is easy to face formatting errors. For Subtask 2, although 12 participants did try to submit for the competition, only 4 managed to submit predictions of the right format. A closer look at the submission files suggests that most of the time, these participants intended to upload predictions for Subtask 1. However, because the default Codalab tab falls on Subtask 2, they make submissions to the wrong task. Nevertheless, we are aware of 1 participant who reached out to try and resolve formatting issues and did not manage to resubmit their predictions in the right format in time. This team ran into issues trying to match the spacing of the original input text.


\subsection{Error Analysis}

For Subtask 1, we had 13 participants while for Subtask 2, we had 4 participants. For Subtask 1, we counted the number of teams that matched the true labels exactly per example. For Subtask 2, if any predicted span exactly match the true Cause-Effect-Signal span, we considered there to be an accurate count. A histogram per subtask reflecting the accuracy counts are reflected in Figure \ref{fig:errors}.

For Subtask 1, 100 examples were predicted correctly, while 4 examples were predicted wrongly by all participants. There is a total of 52 examples that are challenging, where less than half of the participants were able to get a correct prediction.

For Subtask 2, no examples were predicted correctly by all participants. This is because LTRC's submission was very close to the Random Baseline and had no exactly correct predictions. 6 causal relations were predicted correctly by the remaining three participants. Nevertheless, most examples were predicted wrongly by all participants (i.e., 70 examples received all wrong predictions). Clearly, Subtask 2 is a challenging task and has a lot of room for growth.

\section{Conclusion}
\label{sec:conclusion}
In conclusion, our shared task investigated two important tasks in causal text mining, namely: (1) Causal Event Classification, and (2) Cause-Effect-Signal Span Detection. Our shared task attracted 29 registered participants and 17 active participants who made over 100 submissions on the test set. Based on the 12 description papers received, many novel methods that exceeded our initial baseline were proposed. The best F1 scores achieved for Subtask 1 and 2 were 86.19\% and 54.15\% respectively.

We intend to re-launch this shared task next year with even more data for Subtask 2. Additionally, we will also investigate the challenging examples in Subtask 1 that are predicted wrongly by many teams.


\bibliography{custom}

\begin{thebibliography}{63}
\expandafter\ifx\csname natexlab\endcsname\relax\def\natexlab#1{#1}\fi

\bibitem[{Adibhatla and Shrivastava(2022)}]{10}
Hiranmai~Sri Adibhatla and Manish Shrivastava. 2022.
\newblock {{LTRC @ Causal News Corpus 2022}}: Extracting and identifying causal
  elements using adapters.
\newblock In \emph{Proceedings of the 5th Workshop on Challenges and
  Applications of Automated Extraction of Socio-political Events from Text
  (CASE 2022)}, Online. Association for Computational Linguistics.

\bibitem[{Asghar(2016)}]{asghar2016automatic}
Nabiha Asghar. 2016.
\newblock Automatic extraction of causal relations from natural language texts:
  a comprehensive survey.
\newblock \emph{arXiv preprint arXiv:1605.07895}.

\bibitem[{Aziz et~al.(2022)Aziz, Hossain, and Chy}]{23}
Abdul Aziz, Md.~Akram Hossain, and Abu~Nowshed Chy. 2022.
\newblock {{CSECU-DSG @ Causal News Corpus 2022}}: Fusion of {RoBERTa}
  transformer variants for causal event classification.
\newblock In \emph{Proceedings of the 5th Workshop on Challenges and
  Applications of Automated Extraction of Socio-political Events from Text
  (CASE 2022)}, Online. Association for Computational Linguistics.

\bibitem[{Barik et~al.(2016)Barik, Marsi, and
  {\"{O}}zt{\"{u}}rk}]{DBLP:journals/rcs/BarikMO16}
Biswanath Barik, Erwin Marsi, and Pinar {\"{O}}zt{\"{u}}rk. 2016.
\newblock \href
  {http://rcs.cic.ipn.mx/2016\_117/Event\%20Causality\%20Extraction\%20from\%20Natural\%20Science\%20Literature.pdf}
  {Event causality extraction from natural science literature}.
\newblock \emph{Res. Comput. Sci.}, 117:97--107.

\bibitem[{Burdisso et~al.(2022)Burdisso, Zuluaga-Gomez, Fajcik,
  Villatoro-Tello, Singh, Motlicek, and Smrz}]{13}
Sergio Burdisso, Juan Zuluaga-Gomez, Martin Fajcik, Esaú Villatoro-Tello,
  Muskaan Singh, Petr Motlicek, and Pavel Smrz. 2022.
\newblock {{IDIAPers @ Causal News Corpus 2022}}: Causal relation
  identification using a few-shot and prompt-based fine-tuning of language
  models.
\newblock In \emph{The 5th Workshop on Challenges and Applications of Automated
  Extraction of Socio-political Events from Text (CASE @ EMNLP 2022)}, Online.
  Association for Computational Linguistics.

\bibitem[{Cao et~al.(2021)Cao, Zuo, Chen, Liu, Zhao, Chen, and
  Peng}]{cao-etal-2021-knowledge}
Pengfei Cao, Xinyu Zuo, Yubo Chen, Kang Liu, Jun Zhao, Yuguang Chen, and Weihua
  Peng. 2021.
\newblock \href {https://doi.org/10.18653/v1/2021.acl-long.376}
  {Knowledge-enriched event causality identification via latent structure
  induction networks}.
\newblock In \emph{Proceedings of the 59th Annual Meeting of the Association
  for Computational Linguistics and the 11th International Joint Conference on
  Natural Language Processing (Volume 1: Long Papers)}, pages 4862--4872,
  Online. Association for Computational Linguistics.

\bibitem[{Caselli and Vossen(2017)}]{caselli-vossen-2017-event}
Tommaso Caselli and Piek Vossen. 2017.
\newblock \href {https://doi.org/10.18653/v1/W17-2711} {The event {S}tory{L}ine
  corpus: A new benchmark for causal and temporal relation extraction}.
\newblock In \emph{Proceedings of the Events and Stories in the News Workshop},
  pages 77--86, Vancouver, Canada. Association for Computational Linguistics.

\bibitem[{Chen et~al.(2022)Chen, Zhang, Nik, Li, and Fu}]{18}
Xingran Chen, Ge~Zhang, Adam Nik, Mingyu Li, and Jie Fu. 2022.
\newblock {{1Cademy @ Causal News Corpus 2022}}: Enhance causal span detection
  via beam-search-based position selector.
\newblock In \emph{Proceedings of the 5th Workshop on Challenges and
  Applications of Automated Extraction of Socio-political Events from Text
  (CASE 2022)}, Online. Association for Computational Linguistics.

\bibitem[{Dalal et~al.(2021)Dalal, Arcan, and
  Buitelaar}]{dalal-etal-2021-enhancing}
Dhairya Dalal, Mihael Arcan, and Paul Buitelaar. 2021.
\newblock \href {https://doi.org/10.18653/v1/2021.deelio-1.8} {Enhancing
  multiple-choice question answering with causal knowledge}.
\newblock In \emph{Proceedings of Deep Learning Inside Out (DeeLIO): The 2nd
  Workshop on Knowledge Extraction and Integration for Deep Learning
  Architectures}, pages 70--80, Online. Association for Computational
  Linguistics.

\bibitem[{Devlin et~al.(2019)Devlin, Chang, Lee, and
  Toutanova}]{devlin-etal-2019-bert}
Jacob Devlin, Ming-Wei Chang, Kenton Lee, and Kristina Toutanova. 2019.
\newblock \href {https://doi.org/10.18653/v1/N19-1423} {{BERT}: Pre-training of
  deep bidirectional transformers for language understanding}.
\newblock In \emph{Proceedings of the 2019 Conference of the North {A}merican
  Chapter of the Association for Computational Linguistics: Human Language
  Technologies, Volume 1 (Long and Short Papers)}, pages 4171--4186,
  Minneapolis, Minnesota. Association for Computational Linguistics.

\bibitem[{Dunietz et~al.(2020)Dunietz, Burnham, Bharadwaj, Rambow, Chu-Carroll,
  and Ferrucci}]{dunietz-etal-2020-test}
Jesse Dunietz, Greg Burnham, Akash Bharadwaj, Owen Rambow, Jennifer
  Chu-Carroll, and Dave Ferrucci. 2020.
\newblock \href {https://doi.org/10.18653/v1/2020.acl-main.701} {To test
  machine comprehension, start by defining comprehension}.
\newblock In \emph{Proceedings of the 58th Annual Meeting of the Association
  for Computational Linguistics}, pages 7839--7859, Online. Association for
  Computational Linguistics.

\bibitem[{Dunietz et~al.(2017)Dunietz, Levin, and
  Carbonell}]{dunietz-etal-2017-corpus}
Jesse Dunietz, Lori Levin, and Jaime Carbonell. 2017.
\newblock \href {https://doi.org/10.18653/v1/W17-0812} {The {BEC}au{SE} corpus
  2.0: Annotating causality and overlapping relations}.
\newblock In \emph{Proceedings of the 11th Linguistic Annotation Workshop},
  pages 95--104, Valencia, Spain. Association for Computational Linguistics.

\bibitem[{Eckart~de Castilho et~al.(2016)Eckart~de Castilho,
  M{\'u}jdricza-Maydt, Yimam, Hartmann, Gurevych, Frank, and
  Biemann}]{eckart-de-castilho-etal-2016-web}
Richard Eckart~de Castilho, {\'E}va M{\'u}jdricza-Maydt, Seid~Muhie Yimam,
  Silvana Hartmann, Iryna Gurevych, Anette Frank, and Chris Biemann. 2016.
\newblock \href {https://www.aclweb.org/anthology/W16-4011} {A web-based tool
  for the integrated annotation of semantic and syntactic structures}.
\newblock In \emph{Proceedings of the Workshop on Language Technology Resources
  and Tools for Digital Humanities ({LT}4{DH})}, pages 76--84, Osaka, Japan.
  The COLING 2016 Organizing Committee.

\bibitem[{Fajcik et~al.(2022)Fajcik, Singh, Zuluaga-Gomez, Villatoro-Tello,
  Burdisso, Motlicek, and Smrz}]{14}
Martin Fajcik, Muskaan Singh, Juan Zuluaga-Gomez, Esaú Villatoro-Tello, Sergio
  Burdisso, Petr Motlicek, and Pavel Smrz. 2022.
\newblock {{IDIAPers @ Causal News Corpus 2022}}: Extracting
  cause-effect-signal triplets via pre-trained autoregressive language model.
\newblock In \emph{The 5th Workshop on Challenges and Applications of Automated
  Extraction of Socio-political Events from Text (CASE @ EMNLP 2022)}, Online.
  Association for Computational Linguistics.

\bibitem[{Gao et~al.(2019)Gao, Choubey, and Huang}]{gao-etal-2019-modeling}
Lei Gao, Prafulla~Kumar Choubey, and Ruihong Huang. 2019.
\newblock \href {https://doi.org/10.18653/v1/N19-1179} {Modeling document-level
  causal structures for event causal relation identification}.
\newblock In \emph{Proceedings of the 2019 Conference of the North {A}merican
  Chapter of the Association for Computational Linguistics: Human Language
  Technologies, Volume 1 (Long and Short Papers)}, pages 1808--1817,
  Minneapolis, Minnesota. Association for Computational Linguistics.

\bibitem[{Hashimoto et~al.(2014)Hashimoto, Torisawa, Kloetzer, Sano, Varga, Oh,
  and Kidawara}]{hashimoto-etal-2014-toward}
Chikara Hashimoto, Kentaro Torisawa, Julien Kloetzer, Motoki Sano, Istv{\'a}n
  Varga, Jong-Hoon Oh, and Yutaka Kidawara. 2014.
\newblock \href {https://doi.org/10.3115/v1/P14-1093} {Toward future scenario
  generation: Extracting event causality exploiting semantic relation, context,
  and association features}.
\newblock In \emph{Proceedings of the 52nd Annual Meeting of the Association
  for Computational Linguistics (Volume 1: Long Papers)}, pages 987--997,
  Baltimore, Maryland. Association for Computational Linguistics.

\bibitem[{Hassanzadeh et~al.(2019)Hassanzadeh, Bhattacharjya, Feblowitz,
  Srinivas, Perrone, Sohrabi, and Katz}]{ijcai2019-0695}
Oktie Hassanzadeh, Debarun Bhattacharjya, Mark Feblowitz, Kavitha Srinivas,
  Michael Perrone, Shirin Sohrabi, and Michael Katz. 2019.
\newblock \href {https://doi.org/10.24963/ijcai.2019/695} {Answering binary
  causal questions through large-scale text mining: An evaluation using
  cause-effect pairs from human experts}.
\newblock In \emph{Proceedings of the Twenty-Eighth International Joint
  Conference on Artificial Intelligence, {IJCAI-19}}, pages 5003--5009.
  International Joint Conferences on Artificial Intelligence Organization.

\bibitem[{He et~al.(2021)He, Liu, Gao, and Chen}]{DBLP:conf/iclr/HeLGC21}
Pengcheng He, Xiaodong Liu, Jianfeng Gao, and Weizhu Chen. 2021.
\newblock \href {https://openreview.net/forum?id=XPZIaotutsD} {Deberta:
  decoding-enhanced bert with disentangled attention}.
\newblock In \emph{9th International Conference on Learning Representations,
  {ICLR} 2021, Virtual Event, Austria, May 3-7, 2021}. OpenReview.net.

\bibitem[{Hendrickx et~al.(2010)Hendrickx, Kim, Kozareva, Nakov,
  {\'O}~S{\'e}aghdha, Pad{\'o}, Pennacchiotti, Romano, and
  Szpakowicz}]{hendrickx-etal-2010-semeval}
Iris Hendrickx, Su~Nam Kim, Zornitsa Kozareva, Preslav Nakov, Diarmuid
  {\'O}~S{\'e}aghdha, Sebastian Pad{\'o}, Marco Pennacchiotti, Lorenza Romano,
  and Stan Szpakowicz. 2010.
\newblock \href {https://aclanthology.org/S10-1006} {{S}em{E}val-2010 task 8:
  Multi-way classification of semantic relations between pairs of nominals}.
\newblock In \emph{Proceedings of the 5th International Workshop on Semantic
  Evaluation}, pages 33--38, Uppsala, Sweden. Association for Computational
  Linguistics.

\bibitem[{Hidey and McKeown(2016)}]{hidey-mckeown-2016-identifying}
Christopher Hidey and Kathy McKeown. 2016.
\newblock \href {https://doi.org/10.18653/v1/P16-1135} {Identifying causal
  relations using parallel {W}ikipedia articles}.
\newblock In \emph{Proceedings of the 54th Annual Meeting of the Association
  for Computational Linguistics (Volume 1: Long Papers)}, pages 1424--1433,
  Berlin, Germany. Association for Computational Linguistics.

\bibitem[{Hochreiter and Schmidhuber(1997)}]{hochreiter1997long}
Sepp Hochreiter and Jürgen Schmidhuber. 1997.
\newblock \href {https://doi.org/10.1162/neco.1997.9.8.1735} {{Long Short-Term
  Memory}}.
\newblock \emph{Neural Computation}, 9(8):1735--1780.

\bibitem[{H{\"u}rriyeto{\u{g}}lu(2021)}]{case-2021-challenges}
Ali H{\"u}rriyeto{\u{g}}lu, editor. 2021.
\newblock \href {https://aclanthology.org/2021.case-1.0} {\emph{Proceedings of
  the 4th Workshop on Challenges and Applications of Automated Extraction of
  Socio-political Events from Text (CASE 2021)}}. Association for Computational
  Linguistics, Online.

\bibitem[{H{\"u}rriyeto{\u{g}}lu
  et~al.(2021{\natexlab{a}})H{\"u}rriyeto{\u{g}}lu, Mutlu, Y{\"o}r{\"u}k, Liza,
  Kumar, and Ratan}]{hurriyetoglu-etal-2021-multilingual}
Ali H{\"u}rriyeto{\u{g}}lu, Osman Mutlu, Erdem Y{\"o}r{\"u}k, Farhana~Ferdousi
  Liza, Ritesh Kumar, and Shyam Ratan. 2021{\natexlab{a}}.
\newblock \href {https://doi.org/10.18653/v1/2021.case-1.11} {Multilingual
  protest news detection - shared task 1, {CASE} 2021}.
\newblock In \emph{Proceedings of the 4th Workshop on Challenges and
  Applications of Automated Extraction of Socio-political Events from Text
  (CASE 2021)}, pages 79--91, Online. Association for Computational
  Linguistics.

\bibitem[{H{\"u}rriyeto{\u{g}}lu
  et~al.(2021{\natexlab{b}})H{\"u}rriyeto{\u{g}}lu, Tanev, Zavarella,
  Piskorski, Yeniterzi, Mutlu, Yuret, and
  Villavicencio}]{hurriyetoglu-etal-2021-challenges}
Ali H{\"u}rriyeto{\u{g}}lu, Hristo Tanev, Vanni Zavarella, Jakub Piskorski,
  Reyyan Yeniterzi, Osman Mutlu, Deniz Yuret, and Aline Villavicencio.
  2021{\natexlab{b}}.
\newblock \href {https://doi.org/10.18653/v1/2021.case-1.1} {Challenges and
  applications of automated extraction of socio-political events from text
  ({CASE} 2021): Workshop and shared task report}.
\newblock In \emph{Proceedings of the 4th Workshop on Challenges and
  Applications of Automated Extraction of Socio-political Events from Text
  (CASE 2021)}, pages 1--9, Online. Association for Computational Linguistics.

\bibitem[{H{\"u}rriyeto{\u{g}}lu
  et~al.(2020{\natexlab{a}})H{\"u}rriyeto{\u{g}}lu, Y{\"o}r{\"u}k, Zavarella,
  and Tanev}]{aespen-2020-automated}
Ali H{\"u}rriyeto{\u{g}}lu, Erdem Y{\"o}r{\"u}k, Vanni Zavarella, and Hristo
  Tanev, editors. 2020{\natexlab{a}}.
\newblock \href {https://aclanthology.org/2020.aespen-1.0} {\emph{Proceedings
  of the Workshop on Automated Extraction of Socio-political Events from News
  2020}}. European Language Resources Association (ELRA), Marseille, France.

\bibitem[{H{\"u}rriyeto{\u{g}}lu
  et~al.(2020{\natexlab{b}})H{\"u}rriyeto{\u{g}}lu, Zavarella, Tanev,
  Y{\"o}r{\"u}k, Safaya, and Mutlu}]{hurriyetoglu-etal-2020-automated}
Ali H{\"u}rriyeto{\u{g}}lu, Vanni Zavarella, Hristo Tanev, Erdem Y{\"o}r{\"u}k,
  Ali Safaya, and Osman Mutlu. 2020{\natexlab{b}}.
\newblock \href {https://aclanthology.org/2020.aespen-1.1} {Automated
  extraction of socio-political events from news ({AESPEN}): Workshop and
  shared task report}.
\newblock In \emph{Proceedings of the Workshop on Automated Extraction of
  Socio-political Events from News 2020}, pages 1--6, Marseille, France.
  European Language Resources Association (ELRA).

\bibitem[{Hürriyetoğlu et~al.(2021)Hürriyetoğlu, Yörük, Mutlu, Duruşan,
  Yoltar, Yüret, and Gürel}]{10.1162/dint_a_00092}
Ali Hürriyetoğlu, Erdem Yörük, Osman Mutlu, Fırat Duruşan, Çağrı
  Yoltar, Deniz Yüret, and Burak Gürel. 2021.
\newblock \href {https://doi.org/10.1162/dint_a_00092} {{Cross-Context News
  Corpus for Protest Event-Related Knowledge Base Construction}}.
\newblock \emph{Data Intelligence}, 3(2):308--335.

\bibitem[{Izumi et~al.(2021)Izumi, Sano, and Sakaji}]{izumi-etal-2021-economic}
Kiyoshi Izumi, Hitomi Sano, and Hiroki Sakaji. 2021.
\newblock \href {https://aclanthology.org/2021.fnp-1.3} {Economic causal-chain
  search and economic indicator prediction using textual data}.
\newblock In \emph{Proceedings of the 3rd Financial Narrative Processing
  Workshop}, pages 19--25, Lancaster, United Kingdom. Association for
  Computational Linguistics.

\bibitem[{Jo et~al.(2021)Jo, Bang, Reed, and Hovy}]{DBLP:journals/tacl/JoBRH21}
Yohan Jo, Seojin Bang, Chris Reed, and Eduard~H. Hovy. 2021.
\newblock \href {https://transacl.org/ojs/index.php/tacl/article/view/2717}
  {Classifying argumentative relations using logical mechanisms and
  argumentation schemes}.
\newblock \emph{Trans. Assoc. Comput. Linguistics}, 9:721--739.

\bibitem[{Kim et~al.(2022)Kim, Choe, and Lee}]{9}
Juhyeon Kim, Yesong Choe, and Sanghack Lee. 2022.
\newblock {{SNU-Causality Lab @ Causal News Corpus 2022}}: Detecting causality
  by data augmentation via part-of-speech tagging.
\newblock In \emph{Proceedings of the 5th Workshop on Challenges and
  Applications of Automated Extraction of Socio-political Events from Text
  (CASE 2022)}, Online. Association for Computational Linguistics.

\bibitem[{Kohli et~al.(2022)Kohli, Kaur, and Bedi}]{24}
Guneet Kohli, Prabsimran Kaur, and Jatin Bedi. 2022.
\newblock {{ARGUABLY @ Causal News Corpus 2022}}: Contextually augmented
  language models for event causality identification.
\newblock In \emph{Proceedings of the 5th Workshop on Challenges and
  Applications of Automated Extraction of Socio-political Events from Text
  (CASE 2022)}, Online. Association for Computational Linguistics.

\bibitem[{Krumbiegel and Decher(2022)}]{6}
Theresa Krumbiegel and Sophie Decher. 2022.
\newblock {{NLP4ITF @ Causal News Corpus 2022}}: Leveraging linguistic
  information for event causality classification.
\newblock In \emph{Proceedings of the 5th Workshop on Challenges and
  Applications of Automated Extraction of Socio-political Events from Text
  (CASE 2022)}, Online. Association for Computational Linguistics.

\bibitem[{Lee and Sun(2019)}]{DBLP:conf/sigir/LeeS19}
Grace~E. Lee and Aixin Sun. 2019.
\newblock \href {https://doi.org/10.1145/3331184.3331352} {A study on agreement
  in {PICO} span annotations}.
\newblock In \emph{Proceedings of the 42nd International {ACM} {SIGIR}
  Conference on Research and Development in Information Retrieval, {SIGIR}
  2019, Paris, France, July 21-25, 2019}, pages 1149--1152. {ACM}.

\bibitem[{Lewis et~al.(2020)Lewis, Liu, Goyal, Ghazvininejad, Mohamed, Levy,
  Stoyanov, and Zettlemoyer}]{DBLP:conf/acl/LewisLGGMLSZ20}
Mike Lewis, Yinhan Liu, Naman Goyal, Marjan Ghazvininejad, Abdelrahman Mohamed,
  Omer Levy, Veselin Stoyanov, and Luke Zettlemoyer. 2020.
\newblock \href {https://doi.org/10.18653/v1/2020.acl-main.703} {{BART:}
  denoising sequence-to-sequence pre-training for natural language generation,
  translation, and comprehension}.
\newblock In \emph{Proceedings of the 58th Annual Meeting of the Association
  for Computational Linguistics, {ACL} 2020, Online, July 5-10, 2020}, pages
  7871--7880. Association for Computational Linguistics.

\bibitem[{Liu et~al.(2019)Liu, Ott, Goyal, Du, Joshi, Chen, Levy, Lewis,
  Zettlemoyer, and Stoyanov}]{DBLP:journals/corr/abs-1907-11692}
Yinhan Liu, Myle Ott, Naman Goyal, Jingfei Du, Mandar Joshi, Danqi Chen, Omer
  Levy, Mike Lewis, Luke Zettlemoyer, and Veselin Stoyanov. 2019.
\newblock \href {http://arxiv.org/abs/1907.11692} {Roberta: {A} robustly
  optimized {BERT} pretraining approach}.
\newblock \emph{CoRR}, abs/1907.11692.

\bibitem[{Mariko et~al.(2020)Mariko, Abi-Akl, Labidurie, Durfort,
  De~Mazancourt, and El-Haj}]{mariko-etal-2020-financial}
Dominique Mariko, Hanna Abi-Akl, Estelle Labidurie, Stephane Durfort, Hugues
  De~Mazancourt, and Mahmoud El-Haj. 2020.
\newblock \href {https://aclanthology.org/2020.fnp-1.3} {The financial document
  causality detection shared task ({F}in{C}ausal 2020)}.
\newblock In \emph{Proceedings of the 1st Joint Workshop on Financial Narrative
  Processing and MultiLing Financial Summarisation}, pages 23--32, Barcelona,
  Spain (Online). COLING.

\bibitem[{Mariko et~al.(2021)Mariko, Akl, Labidurie, Durfort, de~Mazancourt,
  and El-Haj}]{mariko-etal-2021-financial}
Dominique Mariko, Hanna~Abi Akl, Estelle Labidurie, Stephane Durfort, Hugues
  de~Mazancourt, and Mahmoud El-Haj. 2021.
\newblock \href {https://aclanthology.org/2021.fnp-1.10} {The financial
  document causality detection shared task ({F}in{C}ausal 2021)}.
\newblock In \emph{Proceedings of the 3rd Financial Narrative Processing
  Workshop}, pages 58--60, Lancaster, United Kingdom. Association for
  Computational Linguistics.

\bibitem[{Mirza et~al.(2014)Mirza, Sprugnoli, Tonelli, and
  Speranza}]{mirza-etal-2014-annotating}
Paramita Mirza, Rachele Sprugnoli, Sara Tonelli, and Manuela Speranza. 2014.
\newblock \href {https://doi.org/10.3115/v1/W14-0702} {Annotating causality in
  the {T}emp{E}val-3 corpus}.
\newblock In \emph{Proceedings of the {EACL} 2014 Workshop on Computational
  Approaches to Causality in Language ({CA}to{CL})}, pages 10--19, Gothenburg,
  Sweden. Association for Computational Linguistics.

\bibitem[{Mirza and Tonelli(2014)}]{mirza-tonelli-2014-analysis}
Paramita Mirza and Sara Tonelli. 2014.
\newblock \href {https://aclanthology.org/C14-1198} {An analysis of causality
  between events and its relation to temporal information}.
\newblock In \emph{Proceedings of {COLING} 2014, the 25th International
  Conference on Computational Linguistics: Technical Papers}, pages 2097--2106,
  Dublin, Ireland. Dublin City University and Association for Computational
  Linguistics.

\bibitem[{Nakayama(2018)}]{seqeval}
Hiroki Nakayama. 2018.
\newblock \href {https://github.com/chakki-works/seqeval} {{seqeval}: A python
  framework for sequence labeling evaluation}.
\newblock Software available from https://github.com/chakki-works/seqeval.

\bibitem[{Nguyen and Mitra(2022)}]{15}
Quynh~Anh Nguyen and Arka Mitra. 2022.
\newblock {{NoisyAnnot @ Causal News Corpus 2022}}: Causality detection using
  multiple annotation decision.
\newblock In \emph{Proceedings of the 5th Workshop on Challenges and
  Applications of Automated Extraction of Socio-political Events from Text
  (CASE 2022)}, Online. Association for Computational Linguistics.

\bibitem[{Nik et~al.(2022)Nik, Zhang, Chen, Li, and Fu}]{17}
Adam Nik, Ge~Zhang, Xingran Chen, Mingyu Li, and Jie Fu. 2022.
\newblock {{1Cademy @ Causal News Corpus 2022}}: Leveraging self-training in
  causality classification of socio-political event data.
\newblock In \emph{Proceedings of the 5th Workshop on Challenges and
  Applications of Automated Extraction of Socio-political Events from Text
  (CASE 2022)}, Online. Association for Computational Linguistics.

\bibitem[{Pustejovsky et~al.(2003)Pustejovsky, Casta{\~{n}}o, Ingria,
  Saur{\'{\i}}, Gaizauskas, Setzer, Katz, and
  Radev}]{DBLP:conf/ndqa/PustejovskyCISGSKR03}
James Pustejovsky, Jos{\'{e}}~M. Casta{\~{n}}o, Robert Ingria, Roser
  Saur{\'{\i}}, Robert~J. Gaizauskas, Andrea Setzer, Graham Katz, and
  Dragomir~R. Radev. 2003.
\newblock Timeml: Robust specification of event and temporal expressions in
  text.
\newblock In \emph{New Directions in Question Answering, Papers from 2003
  {AAAI} Spring Symposium, Stanford University, Stanford, CA, {USA}}, pages
  28--34. {AAAI} Press.

\bibitem[{Radinsky et~al.(2012)Radinsky, Davidovich, and
  Markovitch}]{DBLP:conf/www/RadinskyDM12}
Kira Radinsky, Sagie Davidovich, and Shaul Markovitch. 2012.
\newblock \href {https://doi.org/10.1145/2187836.2187958} {Learning causality
  for news events prediction}.
\newblock In \emph{Proceedings of the 21st World Wide Web Conference 2012,
  {WWW} 2012, Lyon, France, April 16-20, 2012}, pages 909--918. {ACM}.

\bibitem[{Radinsky and Horvitz(2013)}]{DBLP:conf/wsdm/RadinskyH13}
Kira Radinsky and Eric Horvitz. 2013.
\newblock \href {https://doi.org/10.1145/2433396.2433431} {Mining the web to
  predict future events}.
\newblock In \emph{Sixth {ACM} International Conference on Web Search and Data
  Mining, {WSDM} 2013, Rome, Italy, February 4-8, 2013}, pages 255--264. {ACM}.

\bibitem[{Ramshaw and Marcus(1995)}]{ramshaw-marcus-1995-text}
Lance Ramshaw and Mitch Marcus. 1995.
\newblock \href {https://aclanthology.org/W95-0107} {Text chunking using
  transformation-based learning}.
\newblock In \emph{Third Workshop on Very Large Corpora}.

\bibitem[{Saha et~al.(2022)Saha, Gittens, Ni, Hassanzadeh, Yener, and
  Srinivas}]{22}
Anik Saha, Alex Gittens, Jian Ni, Oktie Hassanzadeh, Bulent Yener, and Kavitha
  Srinivas. 2022.
\newblock {{SPOCK @ Causal News Corpus 2022}}: Cause-effect-signal span
  detection using span-based and sequence tagging models.
\newblock In \emph{Proceedings of the 5th Workshop on Challenges and
  Applications of Automated Extraction of Socio-political Events from Text
  (CASE 2022)}, Online. Association for Computational Linguistics.

\bibitem[{Sanh et~al.(2019)Sanh, Debut, Chaumond, and
  Wolf}]{DBLP:journals/corr/abs-1910-01108}
Victor Sanh, Lysandre Debut, Julien Chaumond, and Thomas Wolf. 2019.
\newblock \href {http://arxiv.org/abs/1910.01108} {Distilbert, a distilled
  version of {BERT:} smaller, faster, cheaper and lighter}.
\newblock \emph{CoRR}, abs/1910.01108.

\bibitem[{Saur{\i} et~al.(2006)Saur{\i}, Littman, Knippen, Gaizauskas, Setzer,
  and Pustejovsky}]{sauri2006timeml}
Roser Saur{\i}, Jessica Littman, Bob Knippen, Robert Gaizauskas, Andrea Setzer,
  and James Pustejovsky. 2006.
\newblock Timeml annotation guidelines version 1.2.1.

\bibitem[{Stasaski et~al.(2021)Stasaski, Rathod, Tu, Xiao, and
  Hearst}]{stasaski-etal-2021-automatically}
Katherine Stasaski, Manav Rathod, Tony Tu, Yunfang Xiao, and Marti~A. Hearst.
  2021.
\newblock \href {https://aclanthology.org/2021.bea-1.17} {Automatically
  generating cause-and-effect questions from passages}.
\newblock In \emph{Proceedings of the 16th Workshop on Innovative Use of NLP
  for Building Educational Applications}, pages 158--170, Online. Association
  for Computational Linguistics.

\bibitem[{Tan et~al.(2021)Tan, Hazarika, Ng, Poria, and
  Zimmermann}]{tan-etal-2021-causal}
Fiona~Anting Tan, Devamanyu Hazarika, See-Kiong Ng, Soujanya Poria, and Roger
  Zimmermann. 2021.
\newblock \href {https://doi.org/10.18653/v1/2021.cinlp-1.1} {Causal
  augmentation for causal sentence classification}.
\newblock In \emph{Proceedings of the First Workshop on Causal Inference and
  NLP}, pages 1--20, Punta Cana, Dominican Republic. Association for
  Computational Linguistics.

\bibitem[{Tan et~al.(2022{\natexlab{a}})Tan, Hürriyetoğlu, Caselli, Oostdijk,
  Nomoto, Hettiarachchi, Ameer, Uca, Liza, and Hu}]{tan-EtAl:2022:LREC}
Fiona~Anting Tan, Ali Hürriyetoğlu, Tommaso Caselli, Nelleke Oostdijk,
  Tadashi Nomoto, Hansi Hettiarachchi, Iqra Ameer, Onur Uca, Farhana~Ferdousi
  Liza, and Tiancheng Hu. 2022{\natexlab{a}}.
\newblock \href {https://aclanthology.org/2022.lrec-1.246} {The causal news
  corpus: Annotating causal relations in event sentences from news}.
\newblock In \emph{Proceedings of the Language Resources and Evaluation
  Conference}, pages 2298--2310, Marseille, France. European Language Resources
  Association.

\bibitem[{Tan et~al.(2022{\natexlab{b}})Tan, Zuo, and Ng}]{unicausal}
Fiona~Anting Tan, Xinyu Zuo, and See-Kiong Ng. 2022{\natexlab{b}}.
\newblock \href {https://doi.org/10.48550/ARXIV.2208.09163} {Unicausal: Unified
  benchmark and model for causal text mining}.

\bibitem[{Tjong Kim~Sang and
  Buchholz(2000)}]{tjong-kim-sang-buchholz-2000-introduction}
Erik~F. Tjong Kim~Sang and Sabine Buchholz. 2000.
\newblock \href {https://aclanthology.org/W00-0726} {Introduction to the
  {C}o{NLL}-2000 shared task chunking}.
\newblock In \emph{Fourth Conference on Computational Natural Language Learning
  and the Second Learning Language in Logic Workshop}.

\bibitem[{Trust et~al.(2022)Trust, Kadusabe, Minghim, Zahran, Milos, Omala, and
  Yonais}]{16}
Paul Trust, Provia Kadusabe, Rosane Minghim, Ahmed Zahran, Evangelos Milos,
  Kizito Omala, and Haseeb Yonais. 2022.
\newblock {{GGNN @ Causal News Corpus 2022}}: Gated graph neural networks for
  causal event classification from social-political news articles.
\newblock In \emph{Proceedings of the 5th Workshop on Challenges and
  Applications of Automated Extraction of Socio-political Events from Text
  (CASE 2022)}, Online. Association for Computational Linguistics.

\bibitem[{Webber et~al.(2019)Webber, Prasad, Lee, and Joshi}]{webber2019penn}
Bonnie Webber, Rashmi Prasad, Alan Lee, and Aravind Joshi. 2019.
\newblock The penn discourse treebank 3.0 annotation manual.
\newblock \emph{Philadelphia, University of Pennsylvania}.

\bibitem[{Wolf et~al.(2020)Wolf, Debut, Sanh, Chaumond, Delangue, Moi, Cistac,
  Rault, Louf, Funtowicz, Davison, Shleifer, von Platen, Ma, Jernite, Plu, Xu,
  Le~Scao, Gugger, Drame, Lhoest, and Rush}]{wolf-etal-2020-transformers}
Thomas Wolf, Lysandre Debut, Victor Sanh, Julien Chaumond, Clement Delangue,
  Anthony Moi, Pierric Cistac, Tim Rault, Remi Louf, Morgan Funtowicz, Joe
  Davison, Sam Shleifer, Patrick von Platen, Clara Ma, Yacine Jernite, Julien
  Plu, Canwen Xu, Teven Le~Scao, Sylvain Gugger, Mariama Drame, Quentin Lhoest,
  and Alexander Rush. 2020.
\newblock \href {https://doi.org/10.18653/v1/2020.emnlp-demos.6} {Transformers:
  State-of-the-art natural language processing}.
\newblock In \emph{Proceedings of the 2020 Conference on Empirical Methods in
  Natural Language Processing: System Demonstrations}, pages 38--45, Online.
  Association for Computational Linguistics.

\bibitem[{Xu et~al.(2020)Xu, Zuo, Liang, and Zuo}]{xu-etal-2020-review}
Jinghang Xu, Wanli Zuo, Shining Liang, and Xianglin Zuo. 2020.
\newblock \href {https://doi.org/10.18653/v1/2020.coling-main.133} {A review of
  dataset and labeling methods for causality extraction}.
\newblock In \emph{Proceedings of the 28th International Conference on
  Computational Linguistics}, pages 1519--1531, Barcelona, Spain (Online).
  International Committee on Computational Linguistics.

\bibitem[{Yang et~al.(2022)Yang, Han, and Poon}]{DBLP:journals/kais/YangHP22}
Jie Yang, Soyeon~Caren Han, and Josiah Poon. 2022.
\newblock \href {https://doi.org/10.1007/s10115-022-01665-w} {A survey on
  extraction of causal relations from natural language text}.
\newblock \emph{Knowl. Inf. Syst.}, 64(5):1161--1186.

\bibitem[{Yörük et~al.(2021)Yörük, Hürriyetoğlu, Duruşan, and Çağrı
  Yoltar}]{doi:10.1177/00027642211021630}
Erdem Yörük, Ali Hürriyetoğlu, Fırat Duruşan, and Çağrı Yoltar. 2021.
\newblock \href {https://doi.org/10.1177/00027642211021630} {Random sampling in
  corpus design: Cross-context generalizability in automated multicountry
  protest event collection}.
\newblock \emph{American Behavioral Scientist}, 0(0):00027642211021630.

\bibitem[{Zuo et~al.(2021{\natexlab{a}})Zuo, Cao, Chen, Liu, Zhao, Peng, and
  Chen}]{zuo-etal-2021-improving}
Xinyu Zuo, Pengfei Cao, Yubo Chen, Kang Liu, Jun Zhao, Weihua Peng, and Yuguang
  Chen. 2021{\natexlab{a}}.
\newblock \href {https://doi.org/10.18653/v1/2021.findings-acl.190} {Improving
  event causality identification via self-supervised representation learning on
  external causal statement}.
\newblock In \emph{Findings of the Association for Computational Linguistics:
  ACL-IJCNLP 2021}, pages 2162--2172, Online. Association for Computational
  Linguistics.

\bibitem[{Zuo et~al.(2021{\natexlab{b}})Zuo, Cao, Chen, Liu, Zhao, Peng, and
  Chen}]{zuo-etal-2021-learnda}
Xinyu Zuo, Pengfei Cao, Yubo Chen, Kang Liu, Jun Zhao, Weihua Peng, and Yuguang
  Chen. 2021{\natexlab{b}}.
\newblock \href {https://doi.org/10.18653/v1/2021.acl-long.276} {{L}earn{DA}:
  Learnable knowledge-guided data augmentation for event causality
  identification}.
\newblock In \emph{Proceedings of the 59th Annual Meeting of the Association
  for Computational Linguistics and the 11th International Joint Conference on
  Natural Language Processing (Volume 1: Long Papers)}, pages 3558--3571,
  Online. Association for Computational Linguistics.

\bibitem[{Zuo et~al.(2020)Zuo, Chen, Liu, and Zhao}]{zuo-etal-2020-knowdis}
Xinyu Zuo, Yubo Chen, Kang Liu, and Jun Zhao. 2020.
\newblock \href {https://doi.org/10.18653/v1/2020.coling-main.135}
  {{K}now{D}is: Knowledge enhanced data augmentation for event causality
  detection via distant supervision}.
\newblock In \emph{Proceedings of the 28th International Conference on
  Computational Linguistics}, pages 1544--1550, Barcelona, Spain (Online).
  International Committee on Computational Linguistics.

\end{thebibliography}
\bibliographystyle{acl_natbib}

\appendix

\section{Appendix}
\label{sec:appendix}

\subsection{Subtask 2 Agreement Score Calculations}
\label{ssec:agreement}
For span annotations, the agreement scores were calculated by taking a weighted average of the subset level agreement scores that takes into account the example counts per subset. 

We split the training plus development set into 8 subsets and the test set into 2 subsets. While conducting the annotations, the agreement scores were evaluated at a subset level so that we can consistently assess the annotators' performance. The subset level scores takes the average scores between each pair of annotators. For example, if there were three annotators (Annotator A, B, and C) for the subset, then we took the average agreement score when comparing (A,B), (B,C) and (A,C) annotator pairs. Each pair was weighted equally.

The annotator pair level scores were computed by taking the average scores across the sentences. Each sentence was weighted equally. 

At the sentence level, agreement scores were obtained by taking the average scores of each causal relation pair. Each causal relation pair was weighted equally.

Since annotators might annotate multiple spans per example, there are many ways to match the annotated relations between two annotators. We approached this conflict by considering every possible combination pair, after which, we retained the match that returned the highest possible sum of EM, OSB and TO scores. If one annotator identified more causal relations than the other, then EM, OSB and TO scores for that relation is automatically zero.

The KAlpha script was an open-source code\footnote{\url{https://github.com/emerging-welfare/kAlpha}}. The other three metrics were coded based on previous work \citep{DBLP:conf/sigir/LeeS19}.

\end{document}